\documentclass[letterpaper, 10 pt, conference]{ieeeconf}  % Comment this line out if you need a4paper

\IEEEoverridecommandlockouts                              % This command is only needed if 
\usepackage{amsmath}
\usepackage{xcolor}
\usepackage{multirow}
\usepackage{amsfonts}
\usepackage{blindtext}
\usepackage{graphicx}

\usepackage[T1]{fontenc}
\usepackage{tabularray}

\usepackage{amsmath}
\usepackage{xcolor}
\usepackage{amsfonts}
\usepackage{blindtext}
\usepackage{graphicx}
\usepackage{graphicx}
\usepackage[T1]{fontenc}
\usepackage{tabularray}

\title{\LARGE \bf
DensePercept-NCSSD: Vision Mamba towards Real-time Dense Visual Perception with Non-Causal State Space Duality
}

\author{Tushar Anand, Advik Sinha, and Abhijit Das%
  \thanks{Machine Intelligence Group, Department of Computer Science \& Information Systems, Birla Institute of Technology and Science, Pilani, Hyderabad Campus, India. Email: \texttt{abhijit.ads@hyderabad.bits-pilani.ac.in}}%
}

\begin{document}

\maketitle
\thispagestyle{empty}
\pagestyle{empty}

%%%%%%%%%%%%%%%%%%%%%%%%%%%%%%%%%%%%%%%%%%%%%%%%%%%%%%%%%%%%%%%%%%%%%%%%%%%%%%%%
\begin{abstract}

In this work, we propose an accurate and real-time optical flow and disparity estimation model by fusing pairwise input images in the proposed non-causal selective state space for dense perception tasks. We propose a non-causal Mamba block-based model that is fast and efficient and aptly manages the constraints present in a real-time applications. Our proposed model reduces inference times while maintaining high accuracy and low GPU usage for optical flow and disparity map generation. The results and analysis, and validation in real-life scenario justify that our proposed model can be used for unified real-time and accurate 3D dense perception estimation tasks. The code, along with the models, can be found at https://github.com/vimstereo/DensePerceptNCSSD

\end{abstract}

%%%%%%%%%%%%%%%%%%%%%%%%%%%%%%%%%%%%%%%%%%%%%%%%%%%%%%%%%%%%%%%%%%%%%%%%%%%%%%%%
\section{INTRODUCTION}

Efficient optical flow and stereo-disparity estimation from dense perception imagery is a significant challenge in several computer vision robotic systems \cite{hamzah2016literature}. Recent research has shifted towards deep learning-based methods, initially the CNN-based approach \cite{hamid2022stereo, lipson2021raft}, followed by transformer-based modeling \cite{xu2022gmflow}. Deep learning approaches demonstrate higher accuracy for dense perception tasks \cite{xu2022gmflow}, but they are not unsuitable for a real-time applications where fast processing is required with a high frame rate and low computational demands. Works such as \cite{chen2022improvement, coarse2fine} have attempted to propose a real-time model for disparity estimation, but the accuracy of such model was poor. In this direction, to dissolve the trade-off between speed and accuracy for the dense correspondence task, \cite{bora2025vimdisparitybridginggapspeed} proposes a novel disparity estimation method based on visual Mamba with low computation overhead for disparity map generation. In addition, a performance measure that can jointly evaluate the inference speed, computation overhead and the accurateness of a disparity map generation model was proposed. %Most works focus heavily on improving accuracy while compromising on efficiency. Specifically, transformer-based based models which are found to be abundant in literature, hold quadratic complexity because of interleaved attention mechanisms. Due to the this mechanism and longer sequence lengths of transformers, they're computationally expensive to train and deploy.

Recent recent literature of Mamba, state space models (SSM) can capture long-range dependencies and are able to benefit from parallel training. Among the current SSMs, the Mamba \cite{gu2023mamba} block was proposed, which can achieve linear time feature computation while maintaining similar benchmarks as Transformers for different language modeling tasks \cite{gu2023mamba}.  Vision Mamba (ViM) \cite{zhu2024vision} was proposed by incorporating the bidirectional SSMs and position embeddings adopting patch-based analysis to adopt Mamba for the vision task. Further,  VisionMamba \cite{mambavision} was proposed as a hybrid architecture that consists of defining a modified Mamba block along with Transformer blocks for more accurate performance than ViM.  Very recently, Dao and Gu \cite{dao2024transformersssmsgeneralizedmodels} proposed a state space duality (SSD) framework that designs a new architecture Mamba-2. The core layer in Mamba-2 is a refinement of Mamba’s selective SSM. Further development was carried out to design the Visual State Space Model (Vmamba) \cite{liu2024vmambavisualstatespace} which introduces a cross-scanning mechanism to mitigate the problem of one-dimensional scanning in Mamba when applied to vision applications. \cite{shi2024vssdvisionmambanoncausal} were the first to propose a non-causal mamba block which employs a multi-scan strategy and relieves the dependencies of token contribution on previous tokens.

The encouraging results of employing Mamba for disparity estimation in ViM-Disparity \cite{bora2025vimdisparitybridginggapspeed} and recent developments in the literature related to Mamba motivates us to explore the Mamba block further for more efficient and real-time unified dense correspondence tasks. Hence, in this work, we conduct an in-depth analysis of different Mamba block and their possible adaption for the real-time and accurate model for dense perception task estimation. Moreover, ViM-Disparity \cite{bora2025vimdisparitybridginggapspeed} was a hybrid model of SSM and transformer-based attention mechanism, which is found to be an important aspect to feature dense correspondence task. Therefore, there is still a need for a model for dense correspondence task that reduce the quadratic complexity of the attention block. Hence, we introduce a lightweight state-space model which can be a suitable choice to replace the quadratic attention mechanism of the transformer block by non-causal linear-based attention, thereby maintaining the performance and real-time execution for unified vision-dense correspondence tasks adopted from VSSD \cite{shi2024vssdvisionmambanoncausal}. 

VSSD is a recent advancement in computer vision that serves as an alternative to Vision Transformers \cite{dosovitskiy2020image}. Recent works in SSMs, such as state space duality (SSD) \cite{dao2024transformersssmsgeneralizedmodels}, establish a theoretical connection between SSMs and attention mechanisms, especially through structured semi-separable matrices. We introduce a novel Mamba block based on VSSD \cite{shi2024vssdvisionmambanoncausal} for unified dense perception tasks of optical flow and disparity estimation. We have modified the VSSD to learn rich feature representations from individual images and, further from stereo images in parallel through the non-causal mechanism, which is similar to the self and cross-attention mechanism, respectively used in the literature \cite{xu2022gmflow}. Further, to capture long-range within the feature, i.e. in order to capture both large and small pixel displacements, a multi-level of correlation is needed. Hence, we passed the features from the proposed Mamba block through a pyramid-based matching technique based on Gated recurrent Unit (GRU).
%To further improve performance, we incorporate a joint learning block. This joint learning strategy between dense perception inputs enhances the model's ability to predict unified dense perception tasks of disparity and optical flow estimation by effectively capturing relationships between the two views, resulting in more accurate estimations.

%Our approach combines the linear attention of SSMs and a multi-level GRU approach similar to \cite{teed2020raft} to deliver an optimal balance between speed, accuracy, and memory efficiency, making it well-suited for real-time applications on resource-constrained platforms. 

The specific contributions of our work are as follows: 
\begin{itemize}

\item Efficient and real-time Mamba-based architecture for dense perception task estimation of flow and disparity estimation.

\item A modified Mamba block DensePercept-NCSSD based on non-causal SSD that facilitates joint learning of image pair features via a visual correspondence for dense perception task.

\item  We perform an extensive benchmarking analysis of state-of-the-art dense perception task estimation techniques, evaluating our model's inference speed, accuracy, and memory efficiency across multiple datasets, demonstrating its suitability for real-time and resource-constrained environments.

\end{itemize}

\section{Previous Works}
Traditional optical flow methods fall into energy-based \cite{largedisp}, pixel-based \cite{Menze2015GCPR}, and feature-based \cite{revaud2015epicflowedgepreservinginterpolationcorrespondences} approaches. Deep learning techniques \cite{huang2022flowformertransformerarchitectureoptical} have since surpassed them, with FlowNet \cite{fischer2015flownetlearningopticalflow} marking a key shift. Most modern methods like RAFT \cite{lipson2021raft}, use convolutional cost volume,  balancing large motion detection with efficiency. Coarse-to-fine \cite{coarse2fine} and iterative refinement methods \cite{iterrefine} aim at the real-time aspect of optical flow.

Stereo disparity estimation has advanced from CNN-based cascades \cite{mayer2016large} and 3D CNNs \cite{chang2018pyramid} to efficient U-Net models \cite{wang2019anytime} and pyramid pooling \cite{yang2019hierarchical}. Vision Transformers (ViT) \cite{dosovitskiy2020image} inspired attention-based methods \cite{xu2022attention, shen2022pcw}, optimizing cost volumes, while Self-Supervised Learning (SSL) \cite{xu2021digging} further improved accuracy. Any-Net \cite{chen2022improvement} combined 2D-3D CNNs for real-time disparity, and iterative refinement \cite{xu2023iterative} enhanced structural consistency.

Previous research in optical flow and dense disparity perception has typically treated these tasks independently, resulting in separate models. Unified models aim to solve multiple perception tasks with a single architecture. Perceiver IO \cite{jaegle2022perceiveriogeneralarchitecture} introduced a transformer-based approach for optical flow and stereo matching. HD3 \cite{yin2019hierarchicaldiscretedistributiondecomposition} tackled both flow and stereo but lacked transferability. Unimatch \cite{xu2022gmflow} was the first to propose a unified dense perception framework with a shared backbone for all tasks.

\noindent Mamba \cite{gu2023mamba} provides a state-space model (SSM) alternative to Transformers, reducing computational complexity from quadratic to linear. Hybrid architectures where SSMs are combined with attention mechanisms, such as Jamba \cite{lieber2024jambahybridtransformermambalanguage} in language modelling and MambaVision \cite{mambavision} in vision, have demonstrated state-of-the-art performance in various task. To tackle the inherent causal nature of SSM and state space duality (SSD) \cite{dao2024transformersssmsgeneralizedmodels} models, VSSD \cite{shi2024vssdvisionmambanoncausal} proposes a non causal variant suited for vision tasks by employing multi-scan strategies and relieving the dependencies of token contribution on previous tokens.

%%%%%%%%%%%%%%%%%%%%%%%%%%%%%%%%%%%%%%%%%%%%%%%%%%%%%%%%%%%%%%%%%%%%%%%%%%%%%%%%

%%%%%%%%%%%%%%%%%%%%%%%%%%%%%%%%%%%%%%%%%%%%%%%%%%%%%%%%%%%%%%%%%%%%%%%%%%%%%%%%

\begin{figure*}[t]
    \centering
    \includegraphics[width=\textwidth,trim=0 100 0 130,clip]{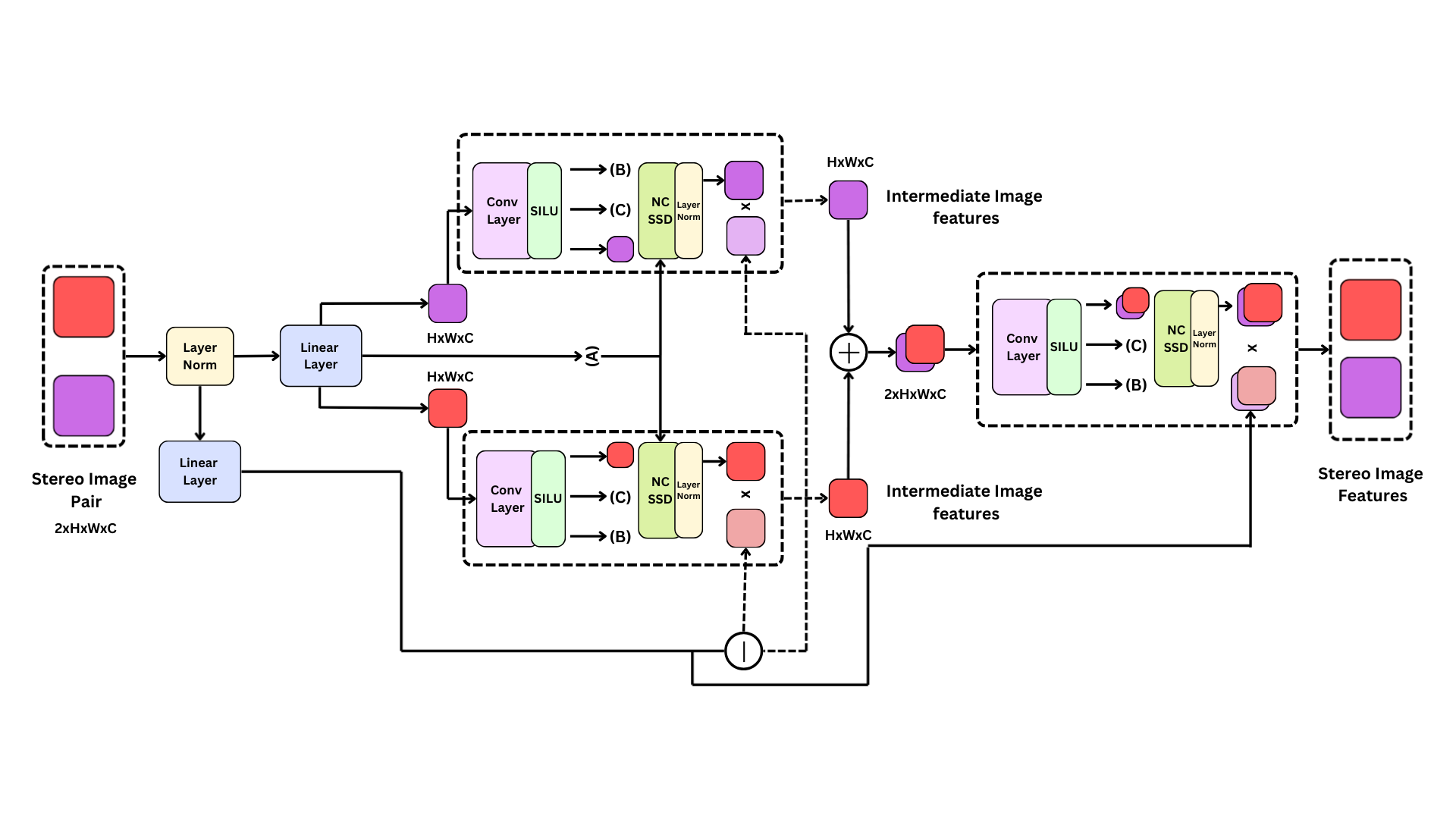}  % Adjust the file name as needed
    \caption{Overview of the proposed DensePercept-NCSSD. The images in the stereo pair are represented in red and purple. (A),(B) and (C) are state matrices. The negative sign(-) represents a split at the batch dimension. The addition sign(+) represents concatenation at the batch dimension.}
    \label{fig:micro_arch}
\end{figure*}

\begin{figure}[t]
    \centering
    \includegraphics[width=\columnwidth]{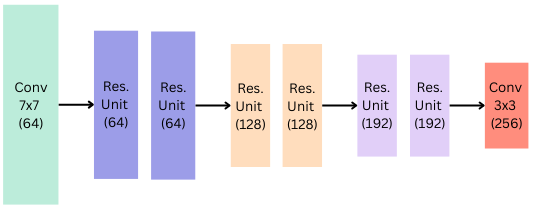}  % Adjust the file name as needed
    \vspace{-2mm}
    \caption{Overview of the context encoder used as reference network.  }
    \label{fig:con}
\end{figure}

\section{Proposed Methodology}
%While transformers have shown strong performance in dense perception tasks, Mamba architectures \cite{bora2025vimdisparitybridginggapspeed} offer a competitive alternative with linear time complexity. To further improve the performance and propose a unified model, we proposed a modified Mamba block based on VSSD and a pyramid-based matching technique to capture the dependency between the pixels for long and short displacement. %We leverage a multi-level GRU-based update mechanism on the cost volume \cite{teed2020raft}, enabling efficient iterative refinement. Inspired by Unimatch, we unify optical flow, stereo matching, and depth estimation into a single, cohesive model.

\subsection{Preliminaries}

\noindent\textbf{Non Causal State Space Duality}: State Space Duality is a special case of selective SSMs that can be implemented in both quadratic and linear forms. The matrix transformations of selective SSMs can be represented as follows:

\begin{align}
y(t) &= \sum_{i=1}^{t} \mathbf{C}_{t}^T \mathbf{A}_{t:i+1} \mathbf{B}_{i} x_{i}, \text{where } \mathbf{A}_{t:i} = \prod_{i=2}^{t} \mathbf{A}_{i} \\
y &= \text{SSM}(\mathbf{A}, \mathbf{B}, \mathbf{C})(x) = \mathbf{F}x, \text{where } \mathbf{F}_{ji} = \mathbf{C}_{j}^T \mathbf{A}_{j:i} \mathbf{B}_{i}
\end{align}

Mamba2 \cite{dao2024transformersssmsgeneralizedmodels} recently simplified the matrix A into a scalar. When A\textsubscript{i} is reduced to a scalar, the linear formula is as follows:
\begin{align}
\centering
h(t) = \mathbf{A}_t h(t - 1) + \mathbf{B}_t x(t), \quad y(t) = \mathbf{C}_t h(t).
\end{align} 

and the quadratic form becomes:

% \begin{align}
% \centering
% \small
% y = F x = \mathbf{M} \cdot \left( \mathbf{C}^T \mathbf{B} \right) x, \quad\text{where } M_{ij} = \begin{cases} 
% A_{i+1} \times \cdots \times A_j & i > j \\ 
% 1 & i = j \\ 
% 0 & i < j 
% \end{cases}
% \end{align}
\begin{align}
y = \mathbf{F} x = \mathbf{M} \cdot \left( \mathbf{C}^T \mathbf{B} \right) x,
\end{align}

\begin{align}
\text{where } M_{ij} =
\begin{cases} 
A_{i+1} \times \cdots \times A_j & i > j \\ 
1 & i = j \\ 
0 & i < j 
\end{cases}
\end{align}

\noindent The scalar ${A_t}$ 
adjusts the impact of the previous hidden state ${h(t-1)}$ and the information at the current time step. In other words, the current hidden state ${h(t)}$ can be seen as a linear combination of the previous hidden state and the current input, weighted by ${A_t}$ and ${1}$, respectively. Thus, if we ignore the absolute values of these two terms and focus only on their relative weighting, we can rewrite it as:
\begin{align}
\centering
h(t) = h(t - 1) + \frac{1}{A_t} \mathbf{B}_t x(t) = \sum_{i=1}^{t} \frac{1}{A_i} \mathbf{B}_i x(i).
\end{align}
In this scenario, the contribution of a specific token to the current hidden state is determined directly by $\frac{1}{A_i}$, rather than by the cumulative product of multiple coefficients. To facilitate the acquisition of global information and better accommodate non-causal image data, VSSD \cite{shi2024vssdvisionmambanoncausal} starts with a bidirectional scanning approach. VSSD has shown that combining the results from both forward and reverse scanning can be effective for this purpose:
% \begin{align}
% \centering
% \mathbf{H}_i = \sum_{j=1}^{i} \frac{1}{A_j} \mathbf{Z}_j + \sum_{j=-L}^{-i} \frac{1}{A_{-j}} \mathbf{Z}_{-j} = \sum_{j=1}^{L} \frac{1}{A_j} \mathbf{Z}_j + \frac{1}{A_i} \mathbf{Z}_i, \text{where } \mathbf{Z}_j = \mathbf{B}_j x(j).
% \end{align}

\begin{align}
\mathbf{H}_i &= \sum_{j=1}^{i} \frac{1}{A_j} \mathbf{Z}_j + \sum_{j=-L}^{-i} \frac{1}{A_{-j}} \mathbf{Z}_{-j} = \sum_{j=1}^{L} \frac{1}{A_j} \mathbf{Z}_j + \frac{1}{A_i} \mathbf{Z}_i, \notag \\
&\text{where } \mathbf{Z}_j = \mathbf{B}_j x(j).
\end{align}

VSSD \cite{shi2024vssdvisionmambanoncausal} treats \( \frac{1}{A_i} \mathbf{Z}_i \) in this equation as a bias and omits it, the above equation simplifies, resulting in all tokens sharing the same hidden state \( \mathbf{H} = \sum_{j=1}^{L} \frac{1}{A_j} \mathbf{Z}_j \). In this case, the forward and reverse scanning results can be seamlessly combined to establish a global context, effectively removing the causal mask and transitioning to a non-causal format. Although these results are derived from a bidirectional scanning approach, in this non-causal format, different scanning paths yield consistent results, making specific scanning routes for capturing global information unnecessary.

Additionally, as shown in the above equation, the contribution of each token to the current hidden state is independent of its spatial distance. Consequently, transforming a flattened 2D feature map into a 1D sequence does not compromise the original structural relationships. Furthermore, the entire computation process can be performed in parallel, avoiding the recurrent methods previously needed for State Space Models (SSMs), which enhances both training and inference speeds. After revising the iteration rules for the hidden state space, VSSD updates the corresponding tensor contraction algorithm or einsum notation in linear form, in line with the Mamba2 framework \cite{dao2024transformersssmsgeneralizedmodels}.
% \begin{align}
% \centering
% \mathbf{Z} = \text{contract}(\text{LD, LN} \rightarrow \text{LND})(\mathbf{X}, \mathbf{B}) \\
% \mathbf{H} = \text{contract}(\text{LL, LDN} \rightarrow \text{ND})(\mathbf{M}, \mathbf{Z}) \\
% \mathbf{Y} = \text{contract}(\text{LN, ND} \rightarrow \text{LD})(\mathbf{C}, \mathbf{H}).
% \end{align}
\begin{align}
\centering
Z &= \text{contract}(\text{LD,LN} \rightarrow \text{LND})(X,B) \\
H &= \text{contract}(\text{LL,LDN} \rightarrow \text{ND})(M,Z) \\
Y &= \text{contract}(\text{LN,ND} \rightarrow \text{LD})(C,H)
\end{align}

This algorithm follows three main steps: first, it expands the input \( X \) using \( B \); second, it unrolls the scalar SSM recurrences to form a global hidden state \( H \); and finally, it contracts \( H \) with \( C \). Compared to the standard SSD, while the operation in the first step remains the same, the sequence length dimension in \( H \) is removed in non-causal mode, as all tokens now share the same hidden state. In the final step, the output \( Y \) is generated by matrix multiplication of \( C \) and \( H \). Since \( M_{i,j} = \frac{1}{A_j} \), the matrix \( M \) can be simplified to a vector \( m \in \mathbb{R}^L \) by removing its first dimension. Integrating \( m \) with either \( X \) or \( B \) in this setup could further simplify the transformation of the equation above.

% \begin{align}
% \centering
% Y &= C(B\textsuperscript{T}(X \cdot m))
% \end{align}

% \begin{figure*}[t]
    
%     \includegraphics[width=0.5\textwidth]{macro_final_arch.png}  % Adjust the file name as needed
%     \caption{Overview of the macro architecture.}
%     \label{fig:macro_arch}
% \end{figure*}

\subsection{Proposed Architecture}

\begin{figure*}[t]
    \centering
    \includegraphics[width=\textwidth]{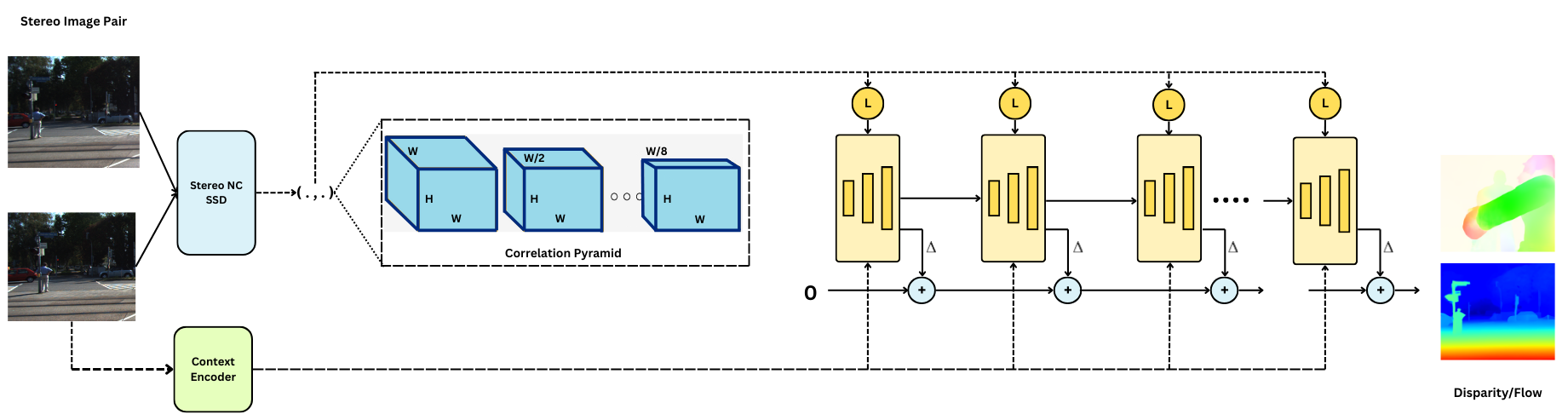}  % Adjust the file name as needed
    \vspace{-8mm}
    \caption{Overview of the macro architecture, which consists of the feature extraction and the machining block.}
    \label{fig:macro_arch}
\end{figure*}

The proposed architecture can broadly be divided into two parts: 1) the feature extraction and 2) the pyramid-based marching (See Fig~\ref{fig:macro_arch}). For feature extraction, we have utilised two separate feature encoders. One set of feature encoding is done via the proposed Mamba block to generate the joint feature from the pair of image (left and right for disparity and consecutive frames for flow). The second encoder acts as the reference for matching. The first set of encoders, the proposed Mamba block DensePercept-NCSSD, obtains dense feature maps at 1/4 resolution and is jointly applied to the left and right images. A detailed description of the Mamba block is given in Figure~\ref{fig:micro_arch}. We also employ a separate context network, as shown in Figure~\ref{fig:con}, with residual blocks and downsampling layers to produce feature maps at 1/4 of the original resolution, which serve as a primary reference. Next, the matching part consists of the correlation computation that finds the visual similarity of every feature map at each pixel location and correspondence for flow/ disparity estimation at different scales at multiple iterations. Further, the relationship between the visual similarity of multi-scale is fostered by a multi-label iterative update. Now, we proceed to explain each of these components in detail.

% \begin{figure}[t]
%     \centering
%     \includegraphics[width=0.4\columnwidth]{Conv 7x7 ().png}  % Adjust the file name as needed
%     \caption{Overview of the Stereo NC SSD architecture. The images in the stereo pair are  represented in red and purple. (A),(B) and (C) are state matrices. Negative sign(-) sign represents split at batch dimension. Addition sign(+) sign represents concatenation at batch dimension.  }
%     \label{fig:context}
% \end{figure}

\subsubsection{Feature Extraction}
The first stage of feature extraction is done in the proposed DensePercept-NCSSD (See Fig~\ref{fig:micro_arch}); where the left and right concatenated images are passed through a Layer Normalization layer to improve the stability and efficiency of the features in the pair of images. It is then split into two branches: the first branch passes through a linear layer to obtain the state transition matrix(A) and the left and right image patches, and the other branch passes through a linear layer to yield linear projections(Z). The individual patches from the first branch are then passed through a dedicated convolution layer and SiLU. From the SiLU, we obtain three outputs: the control input matrix (B) and the output matrix (C) by passing the activated features through a linear layer, and the other branch is simply the image patch feature ${(X_L/X_R)}$. The patches, A, B, C, are then passed through the NC-SSD module and a Layer-Normalization layer to yield ${(Y_L/Y_R)}$. 

The linear projections are split at the batch dimension ${(Z_L/Z_R)}$, and the cross product is computed. The computed cross product acts as a similar functionality of cross-attention in a transformer block, maintaining linear complexity. Further, the cross products are concatenated in a batch dimension. The concatenated tensor is subsequently used to in a similar fashion as before to obtain A,B,C and passed through the NC-SSD in a bidirectional manner. The output is finally passed through a linear layer to obtain individual left and right images for disparity or features from consecutive frames in flow. The set of feature extraction is done in the context encoder (See Fig~\ref{fig:con}), where the reference frame/ image is passed to the context encoder, which contains a CNN with six residual blocks and downsamples each image to 1/8 resolution with D feature maps.

\subsubsection{Pyramid Structure-based Matching}
 The first step in the matching pipeline is to find the visual correspondence. Hence, we construct a correlation volume of features extracted from the left and right components of the proposed Mamba block, i.e DensePercept-NCSSD.
 
 For optical flow, to compute visual correspondence we compute the similarity between all pairs of pixels from the left($I_{l} \in \mathbf{R}^{H \mathbf{x} W \mathbf{x} D}$) and right($I_{r} \in \mathbf{R}^{H \mathbf{x} W \mathbf{x} D}$) image. For this purpose, the correlation volume $C^{o}_{ijkl}$ is computed using a dot product operation between the feature vectors.
\begin{align}
\centering
C^{o}_{ijkl} = \sum (I_{l})_{ijh} \cdot (I_{r})_{klh}, \quad C^{o}_{ijkl} \in \mathbf{R}^{H \mathbf{x} W \mathbf{x} H \mathbf{x} W}
\end{align}

For disparity the 3D correlation volume computes visual similarity between pixels sharing the same y-coordinate, building a 3D correlation volume.

\begin{align}
\centering
C^{disp}_{ijk} = \sum_{h} f_{ijh} \cdot g_{ikh}, \quad C^{disp}_ijk \in \mathbb{R}^{H \times W \times W}
\end{align}

Further, we use the correlation volumes obtained to generate a 4-level pyramid of correlation volumes, i.e the correlation pyramid by average pooling the last 2 dimensions of the optical flow volume and the last dimension of the disparity volume. For the the $\mathbf{k}^{th}$ step we obtain the following optical flow($\mathbf{C}^{o}_{k}$) and disparity($\mathbf{C}^{disp}_{k}$) correlation volumes:
\begin{align}
\centering
C^{o}_{k}\in \mathbf{R}^{H \mathbf{x} W \mathbf{x} H/2^{k} \mathbf{x} W/2^{k}}, \quad
C^{disp}_{k}\in \mathbf{R}^{H \mathbf{x} W \mathbf{x} W/2^{k}}
\end{align}

Next a lookup operator $L_{C}$, is employed to retrieve features from the correlation pyramid to find the relation between small and large pixel displacement. For disparity, it constructs a 1D grid of integer offsets around the current estimate, this grid is used to index pixels from multiple scales, which are linearly interpolated and then concatenated into a single feature map. 

For optical flow, $L_{C}$ maps of each pixel in the first image to its estimated correspondence in the second, defining a local neighbourhood around this point using L1 distance. This neighbourhood, spanning multiple pyramid levels, is indexed using bilinear sampling, with lower levels capturing a broader spatial context. In both cases, the retrieved values from all levels are combined into a unified feature representation.
% \textbf{Correlation Lookup}: The lookup operator $L_{C}$, inspired by RAFT, retrieves features from the correlation pyramid for both stereo disparity and optical flow estimation. For disparity, it constructs a 1D grid of integer offsets around the current estimate, using linear interpolation to sample values from each pyramid level, which are then concatenated into a single feature map. For optical flow, $L_{C}$ maps each pixel in the first image to its estimated correspondence in the second, defining a local neighborhood around this point using L1 distance. This neighborhood, spanning multiple pyramid levels, is indexed using bilinear sampling, with lower levels capturing a broader spatial context. In both cases, the retrieved values from all levels are combined into a unified feature representation.

Both stereo disparity and optical flow estimation iteratively refine predictions at each level of the pyramid, starting from an initial estimate of zero. At each iteration, the current flow or disparity estimate is used to index the generated correlation pyramid, retrieving correlation features, which are then processed by two convolutional layers. The features are then concatenated with context features and passed onto the recurrent update operator employing GRU, which refines the hidden state and predicts the next update.

For disparity estimation, a multi-resolution update strategy is employed, performing updates at 1/8, 1/16, and 1/32 of the input resolution. This allows a larger receptive field, allowing the model to capture large disparities at the lower resolutions while learning intricate features and small disparities at higher resolutions. The final disparity update and correlation lookup are performed at the highest resolution. Optical flow follows a similar iterative process, retrieving correlation features from the pyramid and processing them before updating the flow estimate.

Both methods output predictions at a lower resolution (1/4 or 1/8 of the input image) and apply convex upsampling to restore full resolution. This upsampling takes a weighted combination of a 3×3 grid of coarse resolution neighbours, with weights predicted through convolutional layers. This structured approach ensures efficient and accurate refinement of disparity and optical flow estimates.

\section{Experiments and Analysis}
%\vspace{-1mm}
We used 4x RTX A6000 (48GB) with AMD EPYC 9124 16-Core Processor for our training. For FPS and memory benchmarking, we utilize a single RTX A6000.
%\noindent In this section we will explain and analyze the results of our model for each task independently. We will perform task-level comparisons with other methods and provide an extensive ablation study for our results. The entire model is implemented in PyTorch. \textit{Further results on various model sizes, such as the number of transformer and VisionMamba blocks or depth and number of heads in the attention block are provided in the supplementary.}

\subsection{Optical Flow}

\subsubsection{Implementation details and Metrics}
We train the flow models on the Sceneflow (Flyingthings, Monka and driving) datasets as per the MemFlow protocol \cite{dong2024memflowopticalflowestimation} for 100k steps, with a batch size of 8, and tested on KITTI15 and Sintel. The primary metric used is end-point-error (EPE), the \(l_2\) distance between estimated and ground truth flow vectors. Further EPE is also reported for motion ranges \(s_{0-10}\), \(s_{10-40}\), and \(s_{40+}\). We also employed the F1-all measure (F1A), which indicates the percentage of predicted flow vectors that deviate significantly from the ground truth flow, exceeding a certain threshold (usually 3 pixels) across all pixels in an image. In addition, the frame per second (FPS), memory required (M) and SOMER introduced in \cite {bora2025vimdisparitybridginggapspeed} are measured to evaluate real-time performance. The SOMER metric, which takes into account the FPS, EPE and natural log of memory (\textit{higher the value better is the result}) to produce a unified metric that can jointly evaluate the inference
speed, computation overhead and the accuratenes to measure the real-timeliness of the algorithm.
\vspace{-2mm}

\begin{table}[htb!]
\tiny

\centering

\caption{Results flow on KITTI15 with comparison to exiting and related works in the literature.}
\vspace{-2.5mm}
\label{flow_results}
\scriptsize
\begin{tblr}{
  width = 1\columnwidth,
  colspec = {Q[340]Q[80]Q[78]Q[73]Q[88]Q[65]Q[99]Q[120]Q[180]},
  vlines,
  hline{1-2,3,4,5,6,7,8,9,10} = {-}{},
  hline{3,5,7} = {2-7}{}
}
\textbf{Method}&\textbf{EPE}&\textbf{F1A}&\textbf{\textit{S\textsubscript{0-10}}}&\textbf{\textit{S\textsubscript{10-40}}}&\textbf{S\textsubscript{40+}}&\textbf{FPS}&\textbf{M}&\textbf{SOMER}\\
RAFT \cite{lipson2021raft}&2.45&7.9&0.43&1.18&5.7&11.7&\textbf{180.51}& 0.91\\
Unimatch~ \cite{xu2022gmflow}&2.25&7.2&0.48&1.1&5.12&33.88&236.58&2.75\\
HD3 \cite{hd3}&1.31&6.5&-&-&-&-&-&-\\
UnDAF \cite{undaf} &-&9.56&-&-&-&-&-&-\\
PerceiverIO \cite{jaegle2022perceiveriogeneralarchitecture}&4.98&5.4&-&-&-&-&-&-\\
ViMDisparity \cite{bora2025vimdisparitybridginggapspeed}&2.73&7.41&0.51&1.13&4.94&32.98&238.54&2.2\\ 
MemFlow \cite{dong2024memflowopticalflowestimation}&3.38&12.8&0.46&1.09&5.3&35.27&241.57&1.90\\ 
Proposed &\textbf{0.54}&\textbf{1.4}&\textbf{0.18}&\textbf{0.45}&\textbf{1.20}&\textbf{42.93}& 196.20&\textbf{15.06} \\

\end{tblr}

\end{table}
\vspace{-4mm}

\begin{table}[htb!]
\tiny
\centering
\caption{Ablation on KITTI15 for flow task.}
\vspace{-2.5mm}
\label{flow_results_abl}
\scriptsize
\begin{tblr}{
  width = 1\columnwidth,
  colspec ={Q[130]Q[45]Q[45]Q[45]Q[45]Q[45]Q[45]Q[65]Q[80]},
  vlines,
  hline{1-2,3,4,5,6} = {-}{},
  hline{3,5} = {2-7}{}
}
\textbf{Method} & \textbf{EPE} & \textbf{F1A} & \textbf{\textit{S\textsubscript{0-10}}} & \textbf{\textit{S\textsubscript{10-40}}} & \textbf{S\textsubscript{40+}}&\textbf{FPS} & \textbf{M}&\textbf{SOMER}\\
VSSD \cite{shi2024vssdvisionmambanoncausal} & 5.581 & 14.14 & 0.207 & 0.95 & 15.58 & 24.76 & 208.41& 0.83\\ 
ViM \cite{zhu2024vision} & 0.76 & 2.57 & 0.24 & 0.58 & 1.73 & 37.36 & \textbf{191.09}&9.35 \\
MamVis \cite{mambavision} &0.59& 1.7& 0.20&0.46& 1.32&36.62&218.23 &11.52\\ \hline \hline
Proposed W/O PM &2.04& 6.95&0.42  &0.87   & 4.79 & 37.83 & 228.17&3.41\\
Proposed & \textbf{0.54} & \textbf{1.43} & \textbf{0.18} & \textbf{0.45} & \textbf {1.20} & \textbf{42.93} & 196.20&\textbf{15.06} \\ \hline
\end{tblr}
\end{table}

\begin{table}[!htp]\centering
\caption{Result of EPE for flow task on Sintel. † represents the method that uses the last frame’s flow prediction as initialization for subsequent refinement, while other methods all use two frames only
}
\label{tab:1}
\scriptsize
\begin{tabular}{|l|r|r|r|r|r|}\hline

\multirow{2}{*}{Method} &\multicolumn{2}{c}{Sintel Clean} \vline &\multicolumn{2}{c}{Sintel Final}\vline \\\cline{2-5}
&matched &unmatched &matched &unmatched \\ \hline
FlowNet2 \cite{ilg2016flownet20evolutionoptical} &1.56 &25.4 &2.75 &30.11 \\\hline
PWC-Net+ \cite{sun2018pwcnetcnnsopticalflow} &1.41 &20.12 &2.25 &23.7 \\\hline
HD3 \cite{yin2019hierarchicaldiscretedistributiondecomposition} &1.62 &30.63 &2.17 &24.99 \\\hline
UnDAF \cite{undaf} & 3.91 & - & 5.08 & \\\hline
VCN \cite{yang2019vcn} &1.11 &16.68 &2.22 &22.24 \\\hline
DICL \cite{wang2020displacementinvariantmatchingcostlearning} &0.97 &16.24 &1.66 &19.44 \\\hline
%RAFT &- &- &- &- \\\hline
RAFT† \cite{lipson2021raft} &0.62 &9.65 &1.41 &14.68 \\\hline
GMA† \cite{jiang2021learningestimatehiddenmotions} &0.58 &7.96 &1.24 &12.5 \\\hline
DIP† \cite{zheng2022dipdeepinversepatchmatch} &0.52 &8.92 &1.28 &15.49 \\\hline
AGFlow† \cite{luo2022learningopticalflowadaptive} &0.56 &8.54 &1.22 &12.64 \\\hline
ViM-Dispartity \cite{bora2025vimdisparitybridginggapspeed} &1.43 &8.9&1.71&12.67 \\\hline
CRAFT† \cite{sui2022craftcrossattentionalflowtransformer} &0.61 &8.2 &1.16 &12.64 \\\hline
PERCEIVER IO \cite{jaegle2022perceiveriogeneralarchitecture} & 1.81& -&2.42 &-\\\hline
FlowFormer \cite{huang2022flowformertransformerarchitectureoptical} &0.41 &7.63 &0.99 &11.37 \\\hline
GMFlowNet \cite{zhao2022globalmatchingoverlappingattention} &0.52 &8.49 &1.27 &13.88 \\\hline
GMFlow \cite{xu2022gmflow} &0.65 &10.56 &1.32 &15.8 \\  \hline
Unimatch \cite{xu2022gmflow} &0.34 &6.68 &1.1 &12.74 \\\hline
Proposed & \textbf{0.28} &\textbf{6.31} &\textbf{0.95} &\textbf{10.78} \\\hline

\end{tabular}
\end{table}

% \begin{table}[htb!]
% \centering
% \caption{Results on KITTI15 for flow task.}
% \vspace{-2.5mm}
% \label{flow_results_single}
% \scriptsize
% \begin{tabular}{lccccccc}
% \hline
% \textbf{Method} & \textbf{EPE} & \textbf{F1-all} & \textbf{\textit{S\textsubscript{0-10}}} & \textbf{\textit{S\textsubscript{10-40}}} & \textbf{\textit{S\textsubscript{40+}}} & \textbf{FPS} & \textbf{Memory} \\
% \hline
% RAFT~ \cite{lipson2021raft} ('20)        & 2.45 & 7.9  & 0.43 & 1.18 & 5.7  & 11.7  & \textbf{180.51} \\
% Unimatch~ \cite{xu2022gmflow} ('23)    & 2.25 & 7.2  & 0.48 & 1.10 & 5.12 & 33.88 & 236.58 \\
% MemFlow~ \cite{dong2024memflow} ('24)    & 3.38 & 12.8 & 0.46 & 1.09 & 5.30 & 35.27 & 241.57 \\
% VSSD~ \cite{shi2024vssd} ('24)          & 5.58 & 14.14& 0.21 & 0.95 & 15.58& 24.76 & 208.41 \\
% MambaVision~ \cite{mambavision} ('24)   & 2.04 & 6.95 & 0.42 & 0.87 & 4.79 & 37.83 & 228.17 \\
% Proposed w/o Fusion                    & 0.94 & 2.88 & \textbf{0.15} & 0.42 & 2.25 & \textbf{39.88} & 244.89 \\
% Proposed                               & \textbf{0.86} & \textbf{2.52} & \textbf{0.15} & \textbf{0.41} & \textbf{2.00} & 39.39 & 251.03 \\
% \hline
% \end{tabular}
% \end{table}

\subsubsection{Results and Analysis}
The proposed model outperforms all the works compared from the literature on the KITTI dataset as shown in Table~\ref{flow_results}. Our model achieves state-of-the-art EPE of 0.54, demonstrating significant improvement over MemFlow (3.38), RAFT (2.45) and Unimatch (2.25). Considering F1-all, the proposed model records the lowest score of 1.43, compared to 7.9 and 7.2 for RAFT and Unimatch, respectively, exhibiting improved reliability in handling challenging optical flow estimation instances. The model also demonstrates exceptional performance in handling various motion ranges, achieving EPE values of 0.18, 0.45 and 1.20 in the low-range (S\textsubscript{0-10}), mid-range (S\textsubscript{10-40}) and large-range categories (S\textsubscript{40+}), respectively, outperforming RAFT (0.43, 1.18, 5.7) and Unimatch (0.48, 1.1, 5.12). Our proposed method also outperforms both MambaVision \cite{mambavision} and VSSD, a non-causal Mamba implementation \cite{shi2024vssdvisionmambanoncausal}. The proposed architecture consistently achieves improved accuracy on KITTI while compared to \cite{bora2025vimdisparitybridginggapspeed} across all motion ranges while maintaining competitive performance across various benchmarks. While considering FPS and SOMER the proposed method has outperformed the state-of-the-art. The proposed model demonstrated a marginally higher memory requirement than RAFT but lower compared to any other methods. Further, from Table~\ref{tab:1}, we can conclude that the proposed model was able to achieve better results on Sintel dataset. From the above discussion, it can be concluded that the proposed model was able to attend reliable and real-time flow estimation better than any existing works in the literature. This demonstrates that the proposed method is more effective than quadratic attention for long video sequences while maintaining accuracy and real-time execution.
%\begin{align}
%\centering
%SOMER = \frac{FPS}{EPE\cdot ln(memory)}
%\end{align}
 % \vspace{-1mm}
\begin{table*}[h]
\scriptsize
\caption{Disparity results across datasets (KITTI15, VKITTI, Sintel). compared with different works in the literature.}
\vspace{-2.5mm}
\centering
\begin{tblr}{
  width = \linewidth,
  colspec = {Q[130]Q[50]Q[50]Q[50]Q[50]Q[50]Q[50]Q[50]Q[50]Q[50]},
  hline{1,2,3,4,5,6,7,8,9} = {-}{},
  vline{1,2,4,6,8,11} = {-}{},
  vline{3,5,7,9,10} = {2-14}{}
}
\textbf{Method} & \SetCell[c=2]{c}{\textbf{Kitti15} \cite{geiger2013vision}} && \SetCell[c=2]{c}{\textbf{Vkitti2} \cite{cabon2020virtual}} && \SetCell[c=2]{c}{\textbf{Sintel} \cite{butler2012naturalistic}} \\
& \textbf{EPE} & \textbf{D1} & \textbf{EPE} & \textbf{D1} & \textbf{EPE} & \textbf{D1} & \textbf{FPS} & \textbf{Memory}&\textbf{SOMER}\\
\textbf{Unimatch} ~ \cite{xu2022gmflow}('23)  & 1.21  & 0.05  & 1.95  & 0.13 & 1.45  & \textbf{0.04} & 48.6 & 231.45&7.37\\
\textbf{IGEV} ~ \cite{xu2023iterative}('23)     & 0.38  & 0.37  & 0.92&5.70 & \textbf{0.32}  & 1.18& 1.85&119.18&1.01 \\
\textbf{RAFT} ~ \cite{teed2020raft}('20)     & 1.08  & 4.95  & 0.92 & 6.33 & 0.45  & 1.31 & 2.82 &\textbf{102.41}&0.56 \\
\textbf{Anynet} ~ \cite{chen2022improvement}('18)   & 10.94 & 1.00  & 88.55 & 0.99  &  88.04 & 0.99 & 36 & 240.71&0.6\\
\textbf{ViM-Disparity}('24) \cite{bora2025vimdisparitybridginggapspeed} &1.21&0.05&1.45&\textbf{0.04}&1.95&0.13&51.53&345&6.41\\
\textbf{Proposed}  & \textbf{0.31}  & \textbf{0.015}  & \textbf{0.21}  & \textbf{0.04}  & 0.42  & 0.08 & \textbf{51.71} & 109.93&\textbf{35.36} \\
\end{tblr}
\vspace{1em}
\label{tab_disparity_results}
\end{table*}

\begin{figure*}
    \centering
    \includegraphics[width=1.2\textwidth,trim=250 200 50 200, clip]{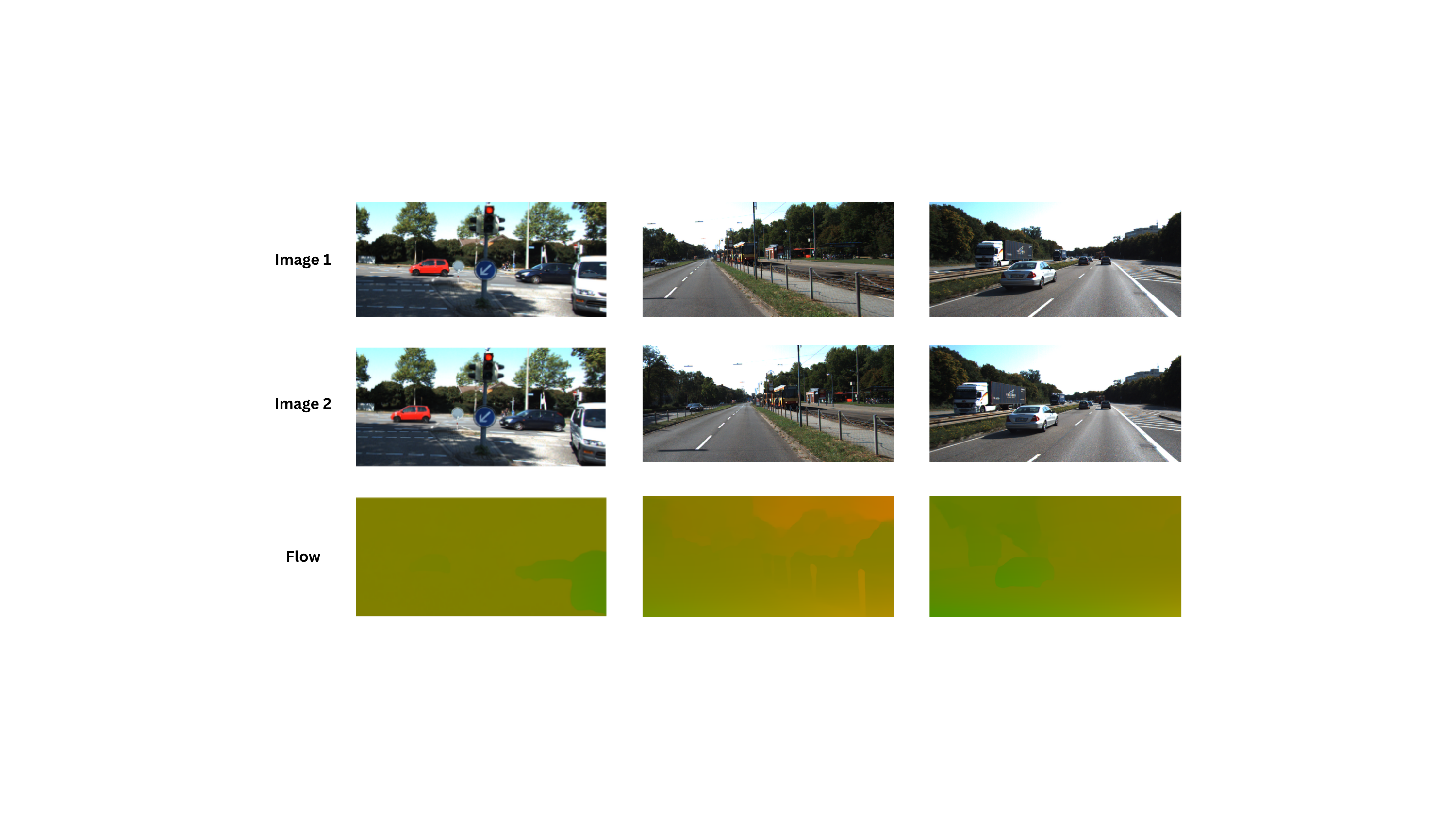}  % Adjust the file name as needed
    \caption{Visual representation of Flow on KITTI15 Dataset }
    \label{fig:context1}
\end{figure*}

For the ablation study, we compared the model with different types of Mamba models available for the vision task, such as Mamba vision (MamVis), Vision Mamba (ViM) and VSSD, and types of matching techniques. It can be observed from Table~\ref{flow_results_abl} combination of pyramid based matching (PM) upsampling and proposed Mamba blocks significantly enhances the model's ability to handle diverse motion magnitudes. The multi-scale nature of the pyramid structure ensures the model can capture both long range and short-range motions. Moreover, in comparison to the existing Mamba block, the proposed Mamba block was able to attend much better results. This proves the effectiveness of the proposed Mamba block. Further, it can be observed that all the components of the proposed model and Mamba block shows a significantly improved FPS compared to others, with memory requirements. Some visual examples of flow estimation on KITTI15 Dataset are in Fig~\ref{fig:context1}.

\begin{figure*}
    \centering
    \includegraphics[width=1.25\textwidth,trim=300 200 75 200, clip]{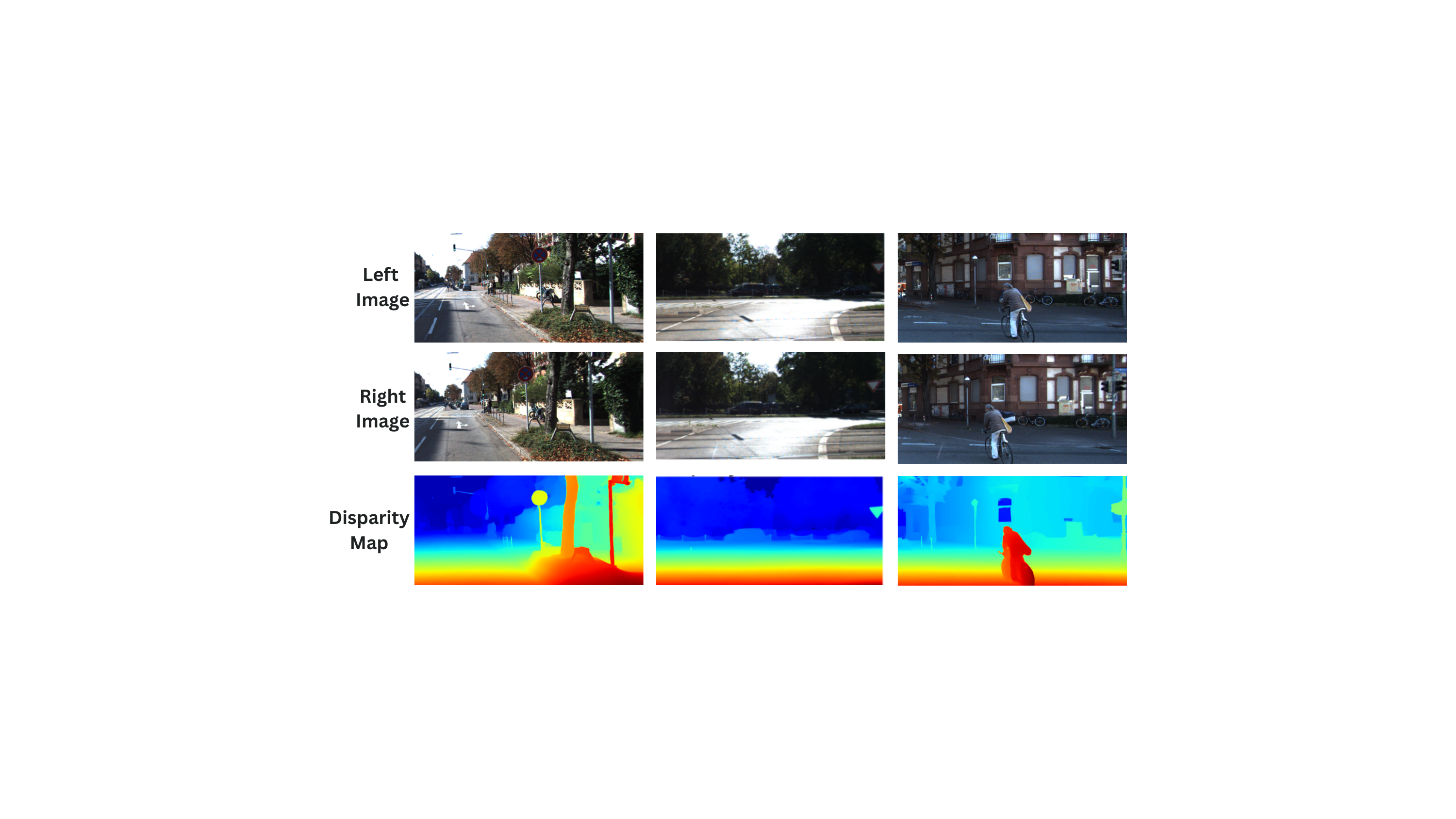}  % Adjust the file name as needed
    \caption{Visual representation of disparity on KITTI15 Dataset }
    \label{fig:context}
\end{figure*}

\subsection{Disparity task}
% \vspace{-3mm}
\subsubsection{Implementation details and Metrics}
For stereo disparity, models are trained on the Sceneflow(Flyingthings, Monka and driving) datasets as per the MemFlow protocol \cite{dong2024memflowopticalflowestimation} dataset for 100k steps with batch size 16 and LR of 2e-4, and tested on KITTI15, VKITTI1 and Sintel. We evaluate performance using common metrics like EPE, D1, FPS, SOMER and memory usage. EPE represents the average L1 distance between predicted and ground truth disparity, whereas D1 indicates the percentage of outliers.

%\vspace{-5mm}
\begin{table*}[]
\scriptsize
\caption{Ablation on disparity task across datasets (KITTI15, VKITTI, Sintel).}
\vspace{-2.5mm}
\centering
\begin{tblr}{
  width = \linewidth,
  colspec = {Q[130]Q[50]Q[50]Q[50]Q[50]Q[50]Q[50]Q[50]Q[50]Q[50]},
  hline{1,2,3,4,5,6,7,8,9} = {-}{},
  vline{1,2,4,6,8,11} = {-}{},
  vline{3,5,7,9,10} = {2-14}{}
}
\textbf{Method} & \SetCell[c=2]{c}{\textbf{Kitti15} \cite{geiger2013vision}} && \SetCell[c=2]{c}{\textbf{Vkitti2} \cite{cabon2020virtual}} && \SetCell[c=2]{c}{\textbf{Sintel} \cite{butler2012naturalistic}} \\
& \textbf{EPE} & \textbf{D1} & \textbf{EPE} & \textbf{D1} & \textbf{EPE} & \textbf{D1} & \textbf{FPS} & \textbf{Memory}&\textbf{SOMER}\\
% \textbf{Unimatch} ~ \cite{xu2022gmflow}('23)  & 1.21  & 0.05  & 1.95  & 0.13 & 1.45  & \textbf{0.04} & 48.6 & 231.45&7.37\\
\textbf{Vision Mamba} ~ \cite{zhu2024vision}('24)   & 1.38 & 0.07  & 1.14 & 0.06  &  11.53 & 0.24 & 51.31 & 334&6.39\\
\textbf{VSSD}('24)  & 0.33  & 0.018  & 0.41  & 0.048  & 0.95  & 0.096 & 33.33 & 281.52&17.91 \\
\textbf{Mamba Vision} ~ \cite{mambavision}('24)   & 0.37 & 0.02  & 0.32 & 0.05  &  0.098 & 0.086 & 49.20 & 141.61&26.84\\
\textbf{Mamba2}('24)  & 0.38  & 0.018  & 0.31  & 0.048  & 0.62  & 0.094 & 50.73&101.48&28.89 \\ \hline \hline
\textbf{Proposed W/O PM}  &0.71&0.02&1.04&0.06&1.43&0.06&\textbf{52.26}&\textbf{98.27}&16.04\\
\textbf{Proposed}  & \textbf{0.31}  & \textbf{0.015}  & \textbf{0.21}  & \textbf{0.041}  & 0.42  & 0.08 & 51.71 & 109.93&\textbf{35.36} \\
\end{tblr}
\vspace{1em}
\label{tab_disparity_ab}
\end{table*}

\subsubsection{Results and Analysis}
The proposed model demonstrates clear improvements in disparity estimation across multiple datasets, particularly in reducing outliers and enhancing efficiency (See Table~\ref{tab_disparity_results}).

On the KITTI15 dataset, it achieves an {EPE of 0.31} and {D1 of 0.015}, significantly outperforming RAFT and Unimatch in outlier reduction. While IGEV attains a similar EPE, its higher D1 (0.37) underscores the importance of balancing accuracy with consistency, which the proposed model handles effectively. On VKITTI, the model records an {EPE of 0.21} and {D1 of 0.041}, again outperforming RAFT and IGEV in outlier percentage, highlighting its robustness in handling complex synthetic datasets. The large error reduction achieved through the combined learning between left and right images in the Stereo NC-SSD blocks and the pyramid based iterative improvements further validate its reliability in difficult scenarios. Similarly, on the extremely challenging Sintel dataset, the model comes in second after IGEV while performing comparably better to Unimatch with an {EPE of 0.42} and {D1 of 0.08}, maintaining a strong balance between accuracy and outlier minimization, even in dynamic scenes. In terms of efficiency, the proposed model delivers superior performance compared to RAFT, Anynet, IGEV and is slightly better than Unimatch (FPS of 48.6), achieving {51.53 FPS}, while maintaining a competitive memory footprint of {109.93} considering the significant improvements in EPE and D1. Considering FPS and SOMER, the proposed model was again best when compared to any work from the literature. Similar to the flow proposed model requires less memory, only RAFT attends a little lower memory requirement than the proposed model. To conclude, the proposed model was able to attend reliable and real-time disparity estimation better than any existing works in the literature. Some visual results of disparity estimation on KITTI15 Dataset are in Fig~\ref{fig:context}.

\textbf{Ablation:} In comparison to VSSD, Mamba2, VisionMamba and MambaVision in the Kitti15, Vkitti2 and Sintel datasets, our method outperformed all scenarios. This proves the effectiveness of the proposed Mamba block for the deparity task. The pyramid based matching is also found to be effective as much better accurateness was found when it is applied, although a little drop in the FPS can be observed. Moreover, the results from Table~\ref{tab_disparity_results} demonstrate that the model is highly suitable for real-time applications, combining speed and scalability with robust accuracy.

%\vspace{-10mm}

\noindent \textbf{Overall observation}: 
The proposed architecture demonstrates robust and consistent performance across multiple tasks, including optical flow and disparity estimation with different conditions (indoor, outdoor, and lighting), which proves its robust performance. By integrating the proposed Mamba block based on non-causal linear-based attention mechanisms from SSDs with traditional pyramid-based refinement mechanisms, the model achieves a significant reduction in error while exhibiting higher FPS and lower memory consumption in comparison to previous works.

\section{Conclusion}
% \vspace{-2.5mm}
\noindent Dense perception task estimation is a well-explored area in the robotic vision community. It is well known that accuracy and real-time generation are found to be a trade-off. Further, most existing techniques aim to enhance the accuracy of the system. This work attempted to extensively analyze the aforementioned trade-off of recent dense perception task estimation methodologies on standard datasets. Concluding from this, we propose a stereo Non-causal SSD model that replaces the quadratic attention of the vision transformer block by linear attention of SSD to bridge the gap of time, memory requirement and accuracy of a real-time dense perception task estimation for flow and disparity. The results and analysis conclude that we could dissolve the speed, accuracy and memory gap with remarkable improvements.

% {
%     \small
%     \bibliographystyle{ieee_fullname.bst}
%     \bibliography{mybibliography}
% }
\bibliographystyle{IEEEtran}


\end{document}